\documentclass[sigconf]{acmart}

\usepackage{soul}
\usepackage{adjustbox}
\usepackage{pgfplots}

\AtBeginDocument{%
  \providecommand\BibTeX{{%
    \normalfont B\kern-0.5em{\scshape i\kern-0.25em b}\kern-0.8em\TeX}}}

\setcopyright{acmcopyright}
\copyrightyear{2021}
\acmYear{2021}
\acmDOI{}

\acmConference[]{}{}{}
\acmBooktitle{}
\acmPrice{}
\acmISBN{}

\begin{document}

\title{RTE: A Tool for Annotating Relation Triplets from Text}

\author{Ankan Mullick}
\authornote{Equal contributions}
\email{ankanm@kgpian.iitkgp.ac.in}
\affiliation{%
 \institution{CSE, IIT Kharagpur}
 \country{India}
}

\author{Animesh Bera}
\email{animesh.connect@gmail.com}
\authornotemark[1]
\affiliation{%
  \institution{Siemens Healthineers}
  \country{India}
}

\author{Tapas Nayak}
\email{tnk02.05@gmail.com}
\authornotemark[1]
\affiliation{%
  \institution{CSE, IIT Kharagpur}
  \country{India}
}







\begin{abstract}
  In this work, we present a Web-based annotation tool `Relation Triplets Extractor' \footnote{https://abera87.github.io/annotate/} (RTE) for annotating relation triplets from the text. Relation extraction is an important task for extracting structured information about real-world entities from the unstructured text available on the Web. In relation extraction, we focus on binary relation that refers to relations between two entities. Recently, many supervised models are proposed to solve this task, but they mostly use noisy training data obtained using the distant supervision method. In many cases, evaluation of the models is also done based on a noisy test dataset. The lack of annotated clean dataset is a key challenge in this area of research. In this work, we built a web-based tool where researchers can annotate datasets for relation extraction on their own very easily. We use a server-less architecture for this tool, and the entire annotation operation is processed using client-side code. Thus it does not suffer from any network latency, and the privacy of the user's data is also maintained. We hope that this tool will be beneficial for the researchers to advance the field of relation extraction.
\end{abstract}

\maketitle

\section{Introduction}

Knowledge bases (KB) are a very useful resource for many downstream applications in natural language understanding such as question-answering, summarization, machine translation, etc. But existing KBs such as Freebase \cite{bollacker2008freebase}, DBpedia \cite{bizer2009dbpedia} have many missing links, and it is difficult to manually complete them. Relation extraction aims to enrich these KBs automatically by extracting entities and relations from the unstructured text available in an abundant amount on the Web. There are two research tracks in the area of relation extraction: (i) Open IE: \cite{banko2007open,etzioni2011reverb,christensen2011srlie,pal2016relnoun} used rule-based systems to identify the verb phrases as relation and noun phrases as entities from text. In this approach relation set is open-ended, and it is referred to as Open IE. (ii) Supervised IE: Here a fixed set of relations are considered. Supervised IE requires a lot of training data of sentence to relation triplets parallel corpus. It is difficult to get such data as existing KBs do not have any associated text with them.

\cite{mintz2009distant,riedel2010modeling,hoffmann2011knowledge} proposed distant supervision to automatically obtain the sentence to triplets parallel corpus to train the supervised models. In distant supervision, if two entities of a relation triplet appear in a sentence, it is selected as a source for that tuple. In this approach, existing relation triplets in the KBs are mapped to text corpora such as Wikipedia articles or New York Times articles. Although this approach can produce a lot of sentence to relation triplet parallel corpus, generally these datasets are noisy. These noisy datasets are used for the training of the models and their evaluation. \cite{mintz2009distant,riedel2010modeling,hoffmann2011knowledge} proposed feature-based learning models, \cite{zeng2014relation,zeng2015distant,huang2016attention,jat2018attention,vashishth2018reside,nayak2019effective} proposed neural network-based models, \cite{bekoulis2018joint,takanobu2019hrlre,nayak2019ptrnetdecoding,Bowen2020JointEO} proposed joint entity and relation extraction models for this task based on such noisy training and test datasets. The lack of annotated data is a major bottleneck in this area of research \cite{Nayak2021DeepNA}. We believe that noisy training data hampers the model's performance negatively. Mechanical turkers from Amazon \footnote{https://www.mturk.com/} can be used for annotating such dataset, but it is costly for many under-funded researchers.

\begin{table}[t]
\small
\centering
\vspace{2mm}
\caption{Example of relation triplets available in text. Our tool provides an easier way to annotate such triplets.}
\vspace{-1mm}
\label{tab:example}
\begin{tabular}{|l|l|}
\hline
Sentence & \begin{tabular}[c]{@{}l@{}}Within a year of that age were Google 's Sergey \\Brin and Larry Page , Apple 's Steve Wozniak ,\\ Yahoo 's Jerry Yang , Skype 's Janus Friis , Chad \\Hurley from YouTube , and Tom Anderson \\ from MySpace .\end{tabular} \\ \hline
Triplets & \begin{tabular}[c]{@{}l@{}}\textless{}Sergey Brin, Google, /business/person/company\textgreater\\ \textless{}Jerry Yang, Yahoo, /business/person/company\textgreater\\ and more ....\end{tabular}\\ \hline
\end{tabular}
\vspace{-1.5mm}
\end{table}

\begin{figure*}[ht]
\caption{Users can add sentences to this page for annotation. Each sentence has a radio button associated with it. Users should select the sentence from the list to annotate entities.}
\vspace{1.5mm}
\label{fig:add_sent}
\centering
\includegraphics[width=1.8\columnwidth]{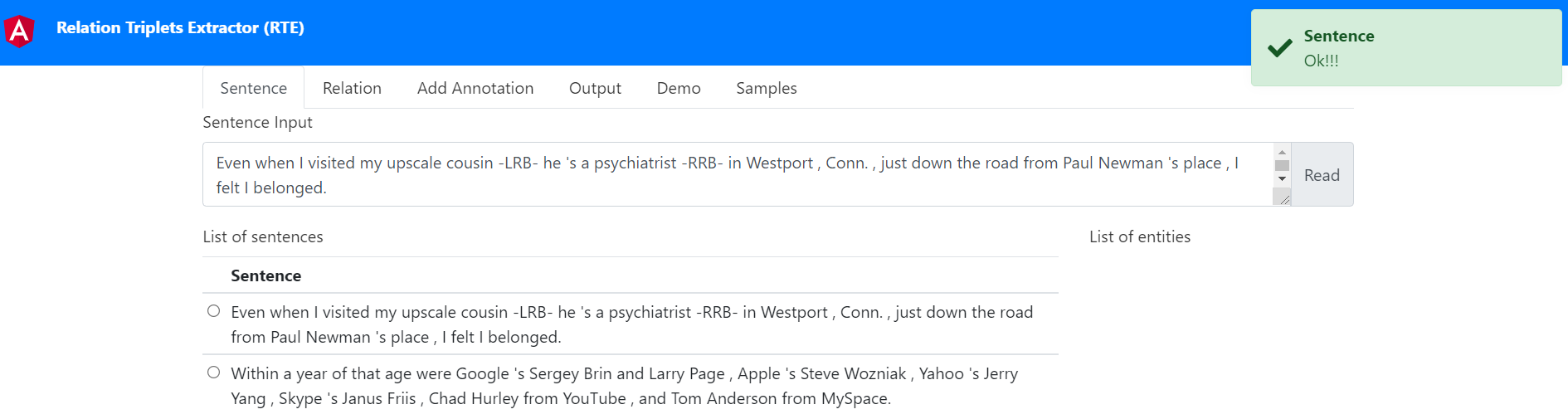}
\end{figure*}

\begin{figure*}[ht]
\vspace{-4mm}
\caption{Users can add relation set in this page.}
\vspace{1.5mm}
\centering
\includegraphics[width=1.8\columnwidth]{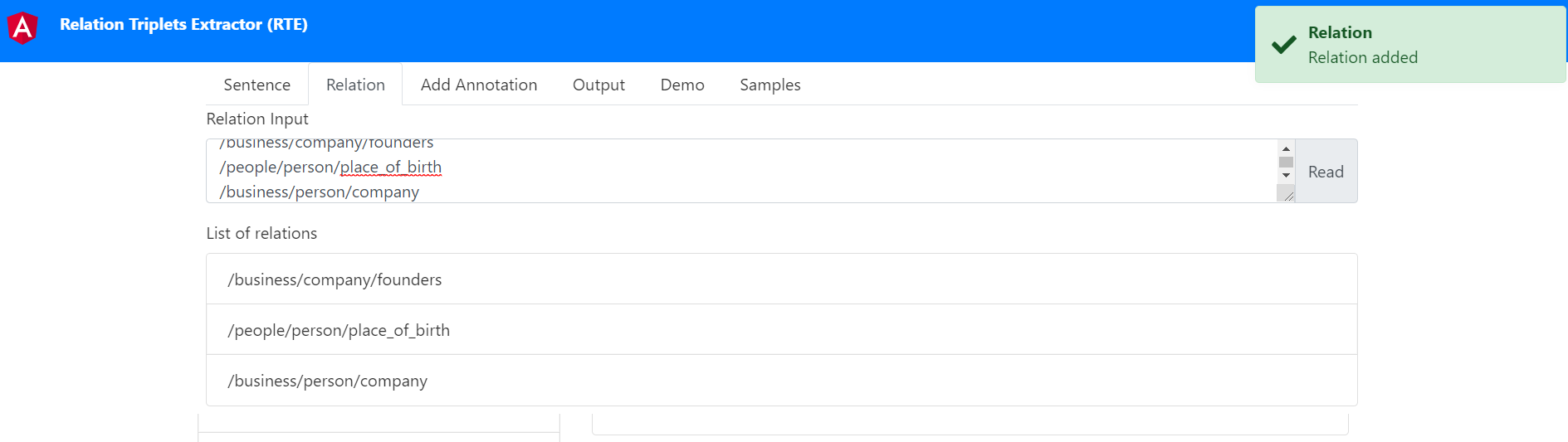}
\label{fig:add_rel}
\vspace{-2mm}
\end{figure*}

In this work, we built a Web-based tool that can be used to annotate datasets for this task. Users can enter sentences and relations in the tool and mark all the entities in them. After that, they can mark the relations among the pairs of entities. This tool doesn't use any server-side code, and the data entered by users is not sent to any remote servers. This ensures that sensitive user data is not leaked during the annotation work. Also, due to server-less architecture, the tool does not face any network latency issues as such. We include a search tool also to search over the relations. We hope that this tool will help the researchers to annotate datasets for their research purposes easily. We added a demo in the tool itself to understand the annotation workflow. We also added few sample sentences and relations in the tool that can be used to familiarize with the annotation workflow.

\section{Methodology}

\subsection{System Design}

This is a web-based tool that can be used to annotate datasets for relation extraction. we used Angular \footnote{https://angular.io/} javascript framework to build this application. This application runs fully on the client-side as a lightweight application. Anybody can extend and customize this tool easily as we develop this module-wise. We have used the following technologies to build this application.\\
\textit{Framework}- Angular 11.0\\
\textit{Language}- TypeScript, JavaScript, HTML5, SCSS\\
\textit{Hosting}- This is a static site and hosted on GitHub \footnote{https://github.com/} page.\\
\textit{Source Control}- GitHub, also is a public repository and any user can download and modify as per their requirement.\\
\textit{Dependencies}- "ang-JSON editor", we used this npm package to beautify and display final triplets output. So that end-user easily copy the output and use their purpose.

We did not use any database to store the user data to protect their data privacy. This is why this tool does not support any partial annotation or resume process in the workflow. We designed it to annotate a small number of instances at one go. The intermediate annotations such as entities and relations are saved in client-side variables, thus refreshing the browser is not allowed. It works as a standalone desktop application. But as it is hosted on the Web, users do not need to install anything to use it. For simplicity, we did not add the file upload/download facility. We opted for the copy/paste mechanism for input/output.

\begin{figure*}[ht]
\caption{Users can annotate the entities present in a sentence. They need to select the entity text and `Add as Entity' option will appear. Annotated entities will appear on the right side. Users can delete an entity from the list too.}
\vspace{2mm}
\centering
\includegraphics[width=1.8\columnwidth]{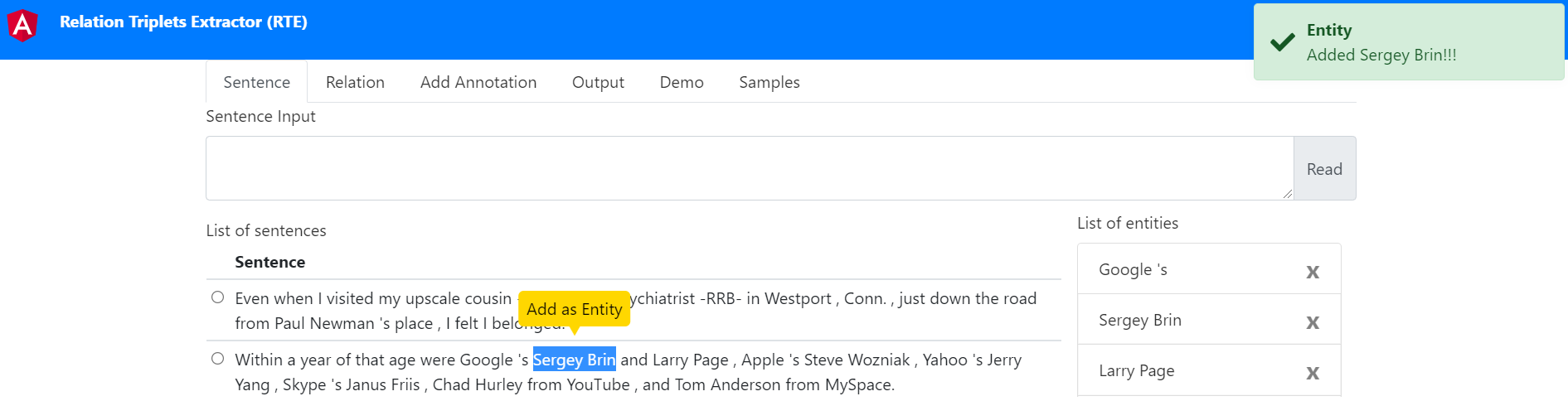}
\label{fig:anno_entity}
\end{figure*}
\vspace{-2mm}
\begin{figure*}[ht]
\caption{Users can annotate the relations between entity pairs on this page. They should select the radio button left of a pair and at the right select the possible relations. There is a search tool for filtering the relations.}
\vspace{2mm}
\centering
\includegraphics[width=1.8\columnwidth]{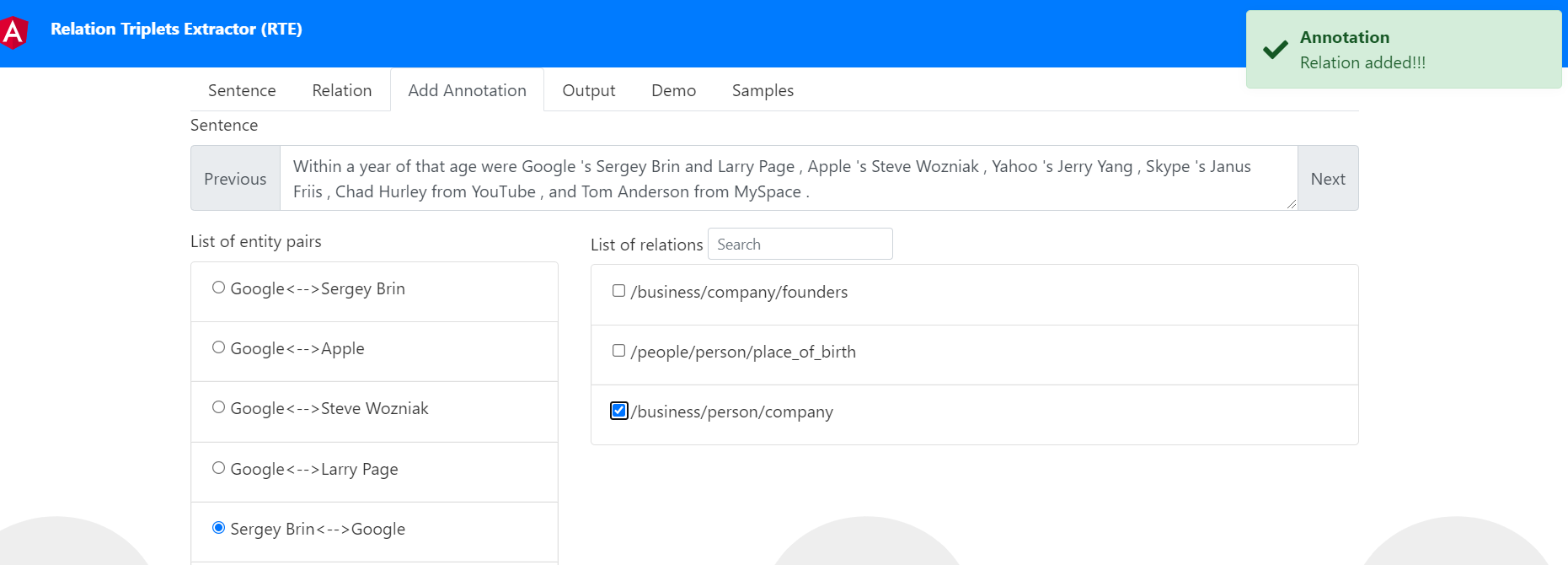}
\label{fig:anno_rel}
\vspace{-1mm}
\end{figure*}

\subsection{Annotation Procedure}

The annotation procedure is very user-friendly on the Github website. The annotation procedure is a four-stage framework - `Sentence Addition', `Relation Set Addition', `Entity Annotation' and `Relation Annotation'. After annotation, the user can view the desired output. 

  
\textbf{Step 1 - Sentence Addition:} In the `Sentence Input' field, the user can provide one or multiple sentences at a time for annotation. These sentences are separated by a newline. After providing the sentences in the `Sentence input' field, a user should push the `Read' button. The list of sentences is shown below and an `Ok!!' message (in green color) is displayed at the right top of the Website. Details are shown in Fig \ref{fig:add_sent}. 

\textbf{Step 2 - Relation Set Addition:} The user needs to provide all possible relations in the `Relation Input' field. Each relation should be in a single line. After adding the relations, the `Read' button is required to be pushed for creating the relation set. The list of relations will be shown below the relation input field. After successful relation set creation, the `relation added' message (in green color) is shown at the right top of the Website. The details are shown in Fig \ref{fig:add_rel}. Some of the relations are added like `/business/company/founders' (relation between company and founders), `/people/person/place\_of\_birth' (relation between person name and place of birth).

\textbf{Step 3 - Entity Annotation:} In the sentence field, the user can select any entity from any sentence, and the `Add as Entity' button comes up in yellow color for adding the entity. After each entity tagging, the added entity is shown at the right top message in green color. A list of entities is shown on the right side of the sentences. The entity tagging is shown in Fig \ref{fig:anno_entity}. The list of entity pairs is shown on the `Add Annotation' page. For example: in the sentence `Within a year of that age were Google 's Sergey Brin and Larry Page, Apple 's Steve Wozniak, Yahoo 's Jerry Yang, Skype 's Janus Friis, Chad Hurley from YouTube, and Tom Anderson from MySpace' following are the entities - `Google', `Sergey Brin', `Larry Page', `Skype', `Janus Friis', `Apple', `Steve Wozniak', `Yahoo', `Jerry Yang', `YouTube', `Chad Hurley', `Tom Anderson', `My Space'. 

\textbf{Step 4 - Relation Annotation:} On the `Add Annotation' page, a list of entity pairs and a list of relations are shown side by side. The user should select an entity pair from the left side, and mark the relations between the two entities from the right side to complete the mapping. When mapping between an entity pair and relation is successful, a `Relation added' message (in green color) is displayed at the right top of the Website. The annotation marking procedure is shown in Fig \ref{fig:anno_rel}. For example: user marks `/business/company/founders' relation to <Google, Sergey Brin> entity pair. Users can easily deselect the relation to remove the relation for any marked entity pair. Users need to deselect the relation checkbox to remove a relation between an entity pair, and data gets saved automatically. 


\noindent \textbf{Output:} After the four stage input procedure the details of the output are shown in a json format in `Output' tab. User can easily collect the annotated outputs for their own purposes. We include the sentence text (\textit{`SentText'}), list of entities (\textit{`EntityMentions'}), list of annotated entity pairs (\textit{`RelationMentions'}) in the output. \textit{`Arg1Text'} and \textit{`Arg2Text'} refers to the first and second entity of the pairs. \textit{`RelationNames'} list contains all the annotated relations for each entity pair. 


\begin{figure*}[ht]
\caption{The output of the annotation is shown in this page in the json format. Users can copy and save accordingly.}
\vspace{2mm}
\centering
\includegraphics[width=1.9\columnwidth]{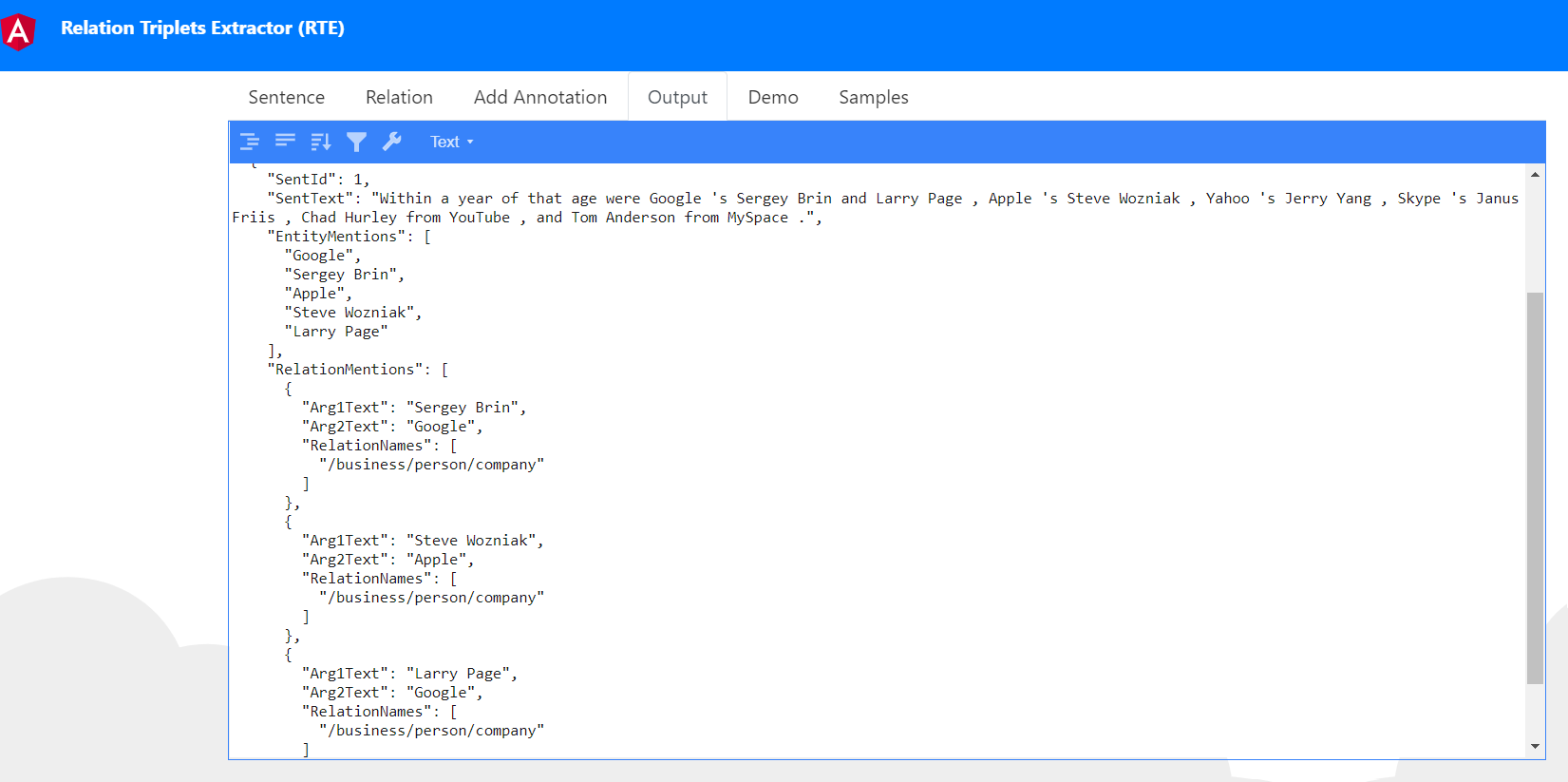}
\label{fig:output}
\end{figure*}

\section{Evaluation \& Comparison}

To evaluate our framework, we compare our annotation tool RTE against WebAnno\footnote{https://webanno.github.io/webanno/}, Brat\footnote{https://github.com/nlplab/brat} and text editor-based annotation to compare the easiness of doing annotation, annotation time (in minutes) per sentence, server space requirements, and set up time. We set up WebAnno and Brat in our server and locally use it for annotation through google chrome. The initial setup time is around 3 hours for WebAnno and 2 hours for Brat. Text editor annotation is done using simple text editor where the user needs to select the entity and save it manually, thereafter add the relations. Two authors annotate the entities and their relations on randomly chosen fifty sentences from Freebase-New York Times dataset \cite{hoffmann2011knowledge}. We measure the total time taken by two authors for fifty sentence annotation (entity and relation) and thereafter calculate the average time (in minute) taken by two authors to annotate one sentence. From Table \ref{tab:comparison}, it is shown that our method (RTE) is comparable to Brat but takes less time for annotation of each sentence and no extra space or initial setup time is required. In the annotated dataset, the average number of entities and relations per sentence are $\sim 6.2$  and $\sim 4.1$. 

\begin{table}[ht]
\centering
\vspace{-2mm}
\caption{Comparison of easiness, average time (in minutes) to annotate each sentence, space requirements and initial set up time needed for different annotation tools.}
\vspace{-2mm}
\label{tab:comparison}
\begin{tabular}{|l|l|l|l|l|}
\hline
Tool        & Text Editor  & WebAnno & Brat & RTE  \\ \hline
Easiness    & Low & Medium  & Good & Good          \\ \hline
Average Time (min)        & 5.76 & 3.12    & 2.53 & 1.87         \\ \hline
Server space & No & Yes     & Yes  & No            \\ \hline
Set up time  & No & Yes     & Yes  & No             \\ \hline
\end{tabular}
\vspace{-3mm}
\end{table}

We show this RTE tool to 10 volunteers to check the effectiveness of the annotation on the same NYT sentences using our tool and competitive tools. In the end, we ask the volunteers to rate the interfaces based on their suitability for this kind of annotation. 7 out of 10 volunteers gave a higher rating to our tool, and the reason they mentioned is the ability to annotate easily and quickly on RTE.



\section{Other Applications}

This tool is useful for many other tasks where the goal is to extract structured data from text. Aspect-Opinion-Sentiment triplet extraction \cite{xu2020position,Jian2021AspectST} is one such task. Users can treat the aspect and opinion spans in the review text as entities and sentiment polarities as relations to annotate data for this task. In the sentence \textit{`Appetizers are excellent ; you can make a great ( but slightly expensive ) meal out of them .'}, users can mark the aspect spans \textit{`Appetizer', `meal'} as entities, the opinion spans \textit{`excellent', `great', `slightly expensive'} as entities too. The sentiment polarities such as \textit{`positive'}, \textit{`negative'}, and \textit{`neutral'} can be added as relations in the `Relation' page. The users need to mark the sentiment for the aspect and opinion span pairs on the `Add Annotation' page of our tool. As example for the pair \textit{(Appetizer, excellent)}, radio button for \textit{`positive'} should be chosen to add the annotation. 

Causality extraction \cite{Dasgupta2018AutomaticEO,Li2021CausalityEB} is another task where this tool is useful for data annotation. The users need to annotate the cause span and effect spans in the text as entities and add the \textit{`Cause-Effect'} as the relation. In the sentence, \textit{`The warmth was radiating from the fireplace to all corners of the room.'}, the spans \textit{`The warmth', `the fireplace'} can be annotated as entities, and \textit{Cause-Effect} relation can be annotated for the pair \textit{(`the fireplace', `The warmth')}.

\vspace{-2mm}
\section{Conclusion}

In this work, we propose a workflow and build a tool to annotate datasets for relation extraction. This tool is easy to use, and researchers can use it to annotate datasets for their experiments. This tool can also be used for annotation for many similar tasks such as aspect extraction, causality extraction. We release the code in Github so that researchers can modify it to suit their tasks as well. We hope that this simple tool will be beneficial for the research community at large.

\bibliographystyle{ACM-Reference-Format}
\bibliography{sample-base}


\begin{thebibliography}{24}


\ifx \showCODEN    \undefined \def \showCODEN     #1{\unskip}     \fi
\ifx \showDOI      \undefined \def \showDOI       #1{#1}\fi
\ifx \showISBNx    \undefined \def \showISBNx     #1{\unskip}     \fi
\ifx \showISBNxiii \undefined \def \showISBNxiii  #1{\unskip}     \fi
\ifx \showISSN     \undefined \def \showISSN      #1{\unskip}     \fi
\ifx \showLCCN     \undefined \def \showLCCN      #1{\unskip}     \fi
\ifx \shownote     \undefined \def \shownote      #1{#1}          \fi
\ifx \showarticletitle \undefined \def \showarticletitle #1{#1}   \fi
\ifx \showURL      \undefined \def \showURL       {\relax}        \fi
\providecommand\bibfield[2]{#2}
\providecommand\bibinfo[2]{#2}
\providecommand\natexlab[1]{#1}
\providecommand\showeprint[2][]{arXiv:#2}

\bibitem[\protect\citeauthoryear{Banko, Cafarella, Soderland, Broadhead, and
  Etzioni}{Banko et~al\mbox{.}}{2007}]%
        {banko2007open}
\bibfield{author}{\bibinfo{person}{Michele Banko}, \bibinfo{person}{Michael~J
  Cafarella}, \bibinfo{person}{Stephen Soderland}, \bibinfo{person}{Matthew
  Broadhead}, {and} \bibinfo{person}{Oren Etzioni}.}
  \bibinfo{year}{2007}\natexlab{}.
\newblock \showarticletitle{Open information extraction from the web.}. In
  \bibinfo{booktitle}{\emph{Proceedings of the International Joint Conference
  on Artificial Intelligence}}.
\newblock


\bibitem[\protect\citeauthoryear{Bekoulis, Deleu, Demeester, and
  Develder}{Bekoulis et~al\mbox{.}}{2018}]%
        {bekoulis2018joint}
\bibfield{author}{\bibinfo{person}{Giannis Bekoulis}, \bibinfo{person}{Johannes
  Deleu}, \bibinfo{person}{Thomas Demeester}, {and} \bibinfo{person}{Chris
  Develder}.} \bibinfo{year}{2018}\natexlab{}.
\newblock \showarticletitle{Joint entity recognition and relation extraction as
  a multi-head selection problem}.
\newblock \bibinfo{journal}{\emph{Expert Systems with Applications}}
  (\bibinfo{year}{2018}).
\newblock


\bibitem[\protect\citeauthoryear{Bizer, Lehmann, Kobilarov, Auer, Becker,
  Cyganiak, and Hellmann}{Bizer et~al\mbox{.}}{2009}]%
        {bizer2009dbpedia}
\bibfield{author}{\bibinfo{person}{Christian Bizer}, \bibinfo{person}{Jens
  Lehmann}, \bibinfo{person}{Georgi Kobilarov}, \bibinfo{person}{S{\"o}ren
  Auer}, \bibinfo{person}{Christian Becker}, \bibinfo{person}{Richard
  Cyganiak}, {and} \bibinfo{person}{Sebastian Hellmann}.}
  \bibinfo{year}{2009}\natexlab{}.
\newblock \showarticletitle{{DB}pedia-{A} Crystallization Point for the Web of
  Data}.
\newblock \bibinfo{journal}{\emph{Web Semantics: Science, Services and Agents
  on the World Wide Web}} (\bibinfo{year}{2009}).
\newblock


\bibitem[\protect\citeauthoryear{Bollacker, Evans, Paritosh, Sturge, and
  Taylor}{Bollacker et~al\mbox{.}}{2008}]%
        {bollacker2008freebase}
\bibfield{author}{\bibinfo{person}{Kurt Bollacker}, \bibinfo{person}{Colin
  Evans}, \bibinfo{person}{Praveen Paritosh}, \bibinfo{person}{Tim Sturge},
  {and} \bibinfo{person}{Jamie Taylor}.} \bibinfo{year}{2008}\natexlab{}.
\newblock \showarticletitle{Freebase: {A} collaboratively created graph
  database for structuring human knowledge}. In
  \bibinfo{booktitle}{\emph{Proceedings of the ACM SIGMOD International
  Conference on Management of Data}}.
\newblock


\bibitem[\protect\citeauthoryear{Bowen, Zhang, Su, Wang, Liu, Wang, and
  Li}{Bowen et~al\mbox{.}}{2020}]%
        {Bowen2020JointEO}
\bibfield{author}{\bibinfo{person}{Yu Bowen}, \bibinfo{person}{Zhenyu Zhang},
  \bibinfo{person}{Jianlin Su}, \bibinfo{person}{Yubin Wang},
  \bibinfo{person}{Tingwen Liu}, \bibinfo{person}{Bin Wang}, {and}
  \bibinfo{person}{Sujian Li}.} \bibinfo{year}{2020}\natexlab{}.
\newblock \showarticletitle{Joint Extraction of Entities and Relations Based on
  a Novel Decomposition Strategy}. In \bibinfo{booktitle}{\emph{ECAI}}.
\newblock


\bibitem[\protect\citeauthoryear{Christensen, Mausam, Soderland, and
  Etzioni}{Christensen et~al\mbox{.}}{2011}]%
        {christensen2011srlie}
\bibfield{author}{\bibinfo{person}{Janara Christensen},
  \bibinfo{person}{Mausam}, \bibinfo{person}{Stephen Soderland}, {and}
  \bibinfo{person}{Oren Etzioni}.} \bibinfo{year}{2011}\natexlab{}.
\newblock \showarticletitle{An analysis of open information extraction based on
  semantic role labeling}. In \bibinfo{booktitle}{\emph{K-CAP}}.
\newblock


\bibitem[\protect\citeauthoryear{Dasgupta, Saha, Dey, and Naskar}{Dasgupta
  et~al\mbox{.}}{2018}]%
        {Dasgupta2018AutomaticEO}
\bibfield{author}{\bibinfo{person}{Tirthankar Dasgupta}, \bibinfo{person}{Rupsa
  Saha}, \bibinfo{person}{Lipika Dey}, {and} \bibinfo{person}{Abir Naskar}.}
  \bibinfo{year}{2018}\natexlab{}.
\newblock \showarticletitle{Automatic Extraction of Causal Relations from Text
  using Linguistically Informed Deep Neural Networks}. In
  \bibinfo{booktitle}{\emph{SIGDIAL Conference}}.
\newblock


\bibitem[\protect\citeauthoryear{Etzioni, Fader, Christensen, Soderland, and
  Mausam}{Etzioni et~al\mbox{.}}{2011}]%
        {etzioni2011reverb}
\bibfield{author}{\bibinfo{person}{Oren Etzioni}, \bibinfo{person}{Anthony
  Fader}, \bibinfo{person}{Janara Christensen}, \bibinfo{person}{Stephen
  Soderland}, {and} \bibinfo{person}{Mausam}.} \bibinfo{year}{2011}\natexlab{}.
\newblock \showarticletitle{Open Information Extraction: The Second
  Generation.}. In \bibinfo{booktitle}{\emph{IJCAI}}.
\newblock


\bibitem[\protect\citeauthoryear{Hoffmann, Zhang, Ling, Zettlemoyer, and
  Weld}{Hoffmann et~al\mbox{.}}{2011}]%
        {hoffmann2011knowledge}
\bibfield{author}{\bibinfo{person}{Raphael Hoffmann}, \bibinfo{person}{Congle
  Zhang}, \bibinfo{person}{Xiao Ling}, \bibinfo{person}{Luke Zettlemoyer},
  {and} \bibinfo{person}{Daniel~S Weld}.} \bibinfo{year}{2011}\natexlab{}.
\newblock \showarticletitle{Knowledge-based weak supervision for information
  extraction of overlapping relations}. In \bibinfo{booktitle}{\emph{ACL}}.
\newblock


\bibitem[\protect\citeauthoryear{Jat, Khandelwal, and Talukdar}{Jat
  et~al\mbox{.}}{2017}]%
        {jat2018attention}
\bibfield{author}{\bibinfo{person}{Sharmistha Jat}, \bibinfo{person}{Siddhesh
  Khandelwal}, {and} \bibinfo{person}{Partha Talukdar}.}
  \bibinfo{year}{2017}\natexlab{}.
\newblock \showarticletitle{Improving Distantly Supervised Relation Extraction
  using Word and Entity Based Attention}. In
  \bibinfo{booktitle}{\emph{Proceedings of the 6th Workshop on Automated
  Knowledge Base Construction}}.
\newblock


\bibitem[\protect\citeauthoryear{Jian, Nayak, Majumder, and Poria}{Jian
  et~al\mbox{.}}{2021}]%
        {Jian2021AspectST}
\bibfield{author}{\bibinfo{person}{Samson Yu~Bai Jian}, \bibinfo{person}{Tapas
  Nayak}, \bibinfo{person}{Navonil Majumder}, {and} \bibinfo{person}{Soujanya
  Poria}.} \bibinfo{year}{2021}\natexlab{}.
\newblock \showarticletitle{Aspect Sentiment Triplet Extraction Using
  Reinforcement Learning}. In \bibinfo{booktitle}{\emph{CIKM}}.
\newblock


\bibitem[\protect\citeauthoryear{Li, Li, Zou, and Ren}{Li
  et~al\mbox{.}}{2021}]%
        {Li2021CausalityEB}
\bibfield{author}{\bibinfo{person}{Zhaoning Li}, \bibinfo{person}{Qi Li},
  \bibinfo{person}{Xiaotian Zou}, {and} \bibinfo{person}{Jiangtao Ren}.}
  \bibinfo{year}{2021}\natexlab{}.
\newblock \showarticletitle{Causality Extraction based on Self-Attentive
  BiLSTM-CRF with Transferred Embeddings}.
\newblock \bibinfo{journal}{\emph{NeuroComputing}} (\bibinfo{year}{2021}).
\newblock


\bibitem[\protect\citeauthoryear{Mintz, Bills, Snow, and Jurafsky}{Mintz
  et~al\mbox{.}}{2009}]%
        {mintz2009distant}
\bibfield{author}{\bibinfo{person}{Mike Mintz}, \bibinfo{person}{Steven Bills},
  \bibinfo{person}{Rion Snow}, {and} \bibinfo{person}{Dan Jurafsky}.}
  \bibinfo{year}{2009}\natexlab{}.
\newblock \showarticletitle{Distant supervision for relation extraction without
  labeled data}. In \bibinfo{booktitle}{\emph{Proceedings of the 47th Annual
  Meeting of the Association for Computational Linguistics and the 4th
  International Joint Conference on Natural Language Processing}}.
\newblock


\bibitem[\protect\citeauthoryear{Nayak, Majumder, Goyal, and Poria}{Nayak
  et~al\mbox{.}}{2021}]%
        {Nayak2021DeepNA}
\bibfield{author}{\bibinfo{person}{Tapas Nayak}, \bibinfo{person}{Navonil
  Majumder}, \bibinfo{person}{Pawan Goyal}, {and} \bibinfo{person}{Soujanya
  Poria}.} \bibinfo{year}{2021}\natexlab{}.
\newblock \showarticletitle{Deep Neural Approaches to Relation Triplets
  Extraction: A Comprehensive Survey}.
\newblock \bibinfo{journal}{\emph{Cognitive Computing}} (\bibinfo{year}{2021}).
\newblock


\bibitem[\protect\citeauthoryear{Nayak and Ng}{Nayak and Ng}{2019}]%
        {nayak2019effective}
\bibfield{author}{\bibinfo{person}{Tapas Nayak} {and} \bibinfo{person}{Hwee~Tou
  Ng}.} \bibinfo{year}{2019}\natexlab{}.
\newblock \showarticletitle{Effective Attention Modeling for Neural Relation
  Extraction}. In \bibinfo{booktitle}{\emph{CoNLL}}.
\newblock


\bibitem[\protect\citeauthoryear{Nayak and Ng}{Nayak and Ng}{2020}]%
        {nayak2019ptrnetdecoding}
\bibfield{author}{\bibinfo{person}{Tapas Nayak} {and} \bibinfo{person}{Hwee~Tou
  Ng}.} \bibinfo{year}{2020}\natexlab{}.
\newblock \showarticletitle{Effective Modeling of Encoder-Decoder Architecture
  for Joint Entity and Relation Extraction}. In
  \bibinfo{booktitle}{\emph{AAAI}}.
\newblock


\bibitem[\protect\citeauthoryear{Pal and Mausam}{Pal and Mausam}{2016}]%
        {pal2016relnoun}
\bibfield{author}{\bibinfo{person}{Harinder Pal} {and}
  \bibinfo{person}{Mausam}.} \bibinfo{year}{2016}\natexlab{}.
\newblock \showarticletitle{Demonyms and Compound Relational Nouns in Nominal
  Open {IE}.}. In \bibinfo{booktitle}{\emph{AKBC}}.
\newblock


\bibitem[\protect\citeauthoryear{Riedel, Yao, and McCallum}{Riedel
  et~al\mbox{.}}{2010}]%
        {riedel2010modeling}
\bibfield{author}{\bibinfo{person}{Sebastian Riedel}, \bibinfo{person}{Limin
  Yao}, {and} \bibinfo{person}{Andrew McCallum}.}
  \bibinfo{year}{2010}\natexlab{}.
\newblock \showarticletitle{Modeling relations and their mentions without
  labeled text}. In \bibinfo{booktitle}{\emph{Proceedings of the European
  Conference on Machine Learning and Knowledge Discovery in Databases}}.
\newblock


\bibitem[\protect\citeauthoryear{Shen and Huang}{Shen and Huang}{2016}]%
        {huang2016attention}
\bibfield{author}{\bibinfo{person}{Yatian Shen} {and} \bibinfo{person}{Xuanjing
  Huang}.} \bibinfo{year}{2016}\natexlab{}.
\newblock \showarticletitle{Attention-based convolutional neural network for
  semantic relation extraction}. In \bibinfo{booktitle}{\emph{Proceedings of
  the 26th International Conference on Computational Linguistics}}.
\newblock


\bibitem[\protect\citeauthoryear{{Takanobu}, {Zhang}, {Liu}, and
  {Huang}}{{Takanobu} et~al\mbox{.}}{2019}]%
        {takanobu2019hrlre}
\bibfield{author}{\bibinfo{person}{Ryuichi {Takanobu}},
  \bibinfo{person}{Tianyang {Zhang}}, \bibinfo{person}{Jiexi {Liu}}, {and}
  \bibinfo{person}{Minlie {Huang}}.} \bibinfo{year}{2019}\natexlab{}.
\newblock \showarticletitle{A Hierarchical Framework for Relation Extraction
  with Reinforcement Learning}. In \bibinfo{booktitle}{\emph{AAAI}}.
\newblock


\bibitem[\protect\citeauthoryear{Vashishth, Joshi, Prayaga, Bhattacharyya, and
  Talukdar}{Vashishth et~al\mbox{.}}{2018}]%
        {vashishth2018reside}
\bibfield{author}{\bibinfo{person}{Shikhar Vashishth}, \bibinfo{person}{Rishabh
  Joshi}, \bibinfo{person}{Sai~Suman Prayaga}, \bibinfo{person}{Chiranjib
  Bhattacharyya}, {and} \bibinfo{person}{Partha Talukdar}.}
  \bibinfo{year}{2018}\natexlab{}.
\newblock \showarticletitle{RESIDE: Improving Distantly-Supervised Neural
  Relation Extraction using Side Information}. In
  \bibinfo{booktitle}{\emph{Proceedings of the 2018 Conference on Empirical
  Methods in Natural Language Processing}}.
\newblock


\bibitem[\protect\citeauthoryear{Xu, Li, Lu, and Bing}{Xu
  et~al\mbox{.}}{2020}]%
        {xu2020position}
\bibfield{author}{\bibinfo{person}{Lu Xu}, \bibinfo{person}{Hao Li},
  \bibinfo{person}{Wei Lu}, {and} \bibinfo{person}{Lidong Bing}.}
  \bibinfo{year}{2020}\natexlab{}.
\newblock \showarticletitle{Position-Aware Tagging for Aspect Sentiment Triplet
  Extraction}. In \bibinfo{booktitle}{\emph{EMNLP}}.
\newblock


\bibitem[\protect\citeauthoryear{Zeng, Liu, Chen, and Zhao}{Zeng
  et~al\mbox{.}}{2015}]%
        {zeng2015distant}
\bibfield{author}{\bibinfo{person}{Daojian Zeng}, \bibinfo{person}{Kang Liu},
  \bibinfo{person}{Yubo Chen}, {and} \bibinfo{person}{Jun Zhao}.}
  \bibinfo{year}{2015}\natexlab{}.
\newblock \showarticletitle{Distant Supervision for Relation Extraction via
  Piecewise Convolutional Neural Networks.}. In
  \bibinfo{booktitle}{\emph{Proceedings of the 2015 Conference on Empirical
  Methods in Natural Language Processing}}.
\newblock


\bibitem[\protect\citeauthoryear{Zeng, Liu, Lai, Zhou, and Zhao}{Zeng
  et~al\mbox{.}}{2014}]%
        {zeng2014relation}
\bibfield{author}{\bibinfo{person}{Daojian Zeng}, \bibinfo{person}{Kang Liu},
  \bibinfo{person}{Siwei Lai}, \bibinfo{person}{Guangyou Zhou}, {and}
  \bibinfo{person}{Jun Zhao}.} \bibinfo{year}{2014}\natexlab{}.
\newblock \showarticletitle{Relation Classification via Convolutional Deep
  Neural Network.}. In \bibinfo{booktitle}{\emph{Proceedings of the 25th
  International Conference on Computational Linguistics}}.
\newblock


\end{thebibliography}

\end{document}